\documentclass[twocolumn]{svjour3}
\usepackage[english]{babel}
\usepackage{subfigure}
\usepackage{amsmath}
\usepackage{graphicx}
\usepackage{subfigure}
\usepackage{listings}

\begin{document}
\title{Using Decision Trees for Interpretable Supervised Clustering}
\author{Natallia Kokash*\thanks{*Corresponding author}
\and Leonid Makhnist}

\institute{N. Kokash \at University of Amsterdam, Kloveniersburgwal 48, 1012 CX Amsterdam, The Netherlands
\email{natallia.kokash@gmail.com}
\and
L. Makhnist \at Brest State Technical University, Moskovskaya
267, 224017 Brest, Republic of Belarus. 
\email{hm@bstu.bu}
}

\authorrunning{N. Kokash, L. Makhnist}
\date{Created: 27-03-2021}

\maketitle

\begin{abstract}
In this paper, we address an issue of finding explainable clusters of class-uniform data in labelled datasets. The issue falls into the domain of interpretable supervised clustering. Unlike traditional clustering, supervised clustering aims at forming clusters of labelled data with high probability densities. We are particularly interested in finding clusters of data of a given class and describing the clusters with the set of comprehensive rules. We propose an iterative method to extract high-density clusters with the help of decision-tree-based classifiers as the most intuitive learning method, and discuss the method of node selection to maximize quality of identified groups.   
\end{abstract}

\lstdefinelanguage{pseudo}{
    basicstyle=\scriptsize\ttfamily,
    stepnumber=1,
    numbersep=8pt,
    showstringspaces=false,
    breaklines=true
}

\section{Introduction}
\label{sect:Introduction}

How would a human characterize passengers surviving the Titanic disaster? In as few words as possible, it is not unreasonable to say:
\begin{itemize}
    \item they are mainly first class passengers (indeed, $61\%$ of passengers in the first class survived as opposed to $42\%$ in the second and $24\%$ in the third classes),
    \item female passengers ($75\%$ of female passengers and only $20\%$ of male passengers escaped perishing),
    \item female crew members had the highest survival rate  ($87\%$ of female crew members successfully evacuated).
\end{itemize}
Concise collective definition of population groups like in the example above generalizes individual observations and enables people to grasp existing trends in otherwise incomprehensible or hard to interpret data. In application areas such as logistics, urban-planning, supply-and-demand management, marketing, healthcare and so on, identifying target groups is essential for adjusting existing policies to the expectations of major groups of stakeholders.  

With the adoption of machine learning methods for decision-making in industries with high accountability demands such as healthcare or financial services, we observed a spurge in interest for model interpretability~\cite{ML-interpret19,Vellido2020}. In the context of our study, we understand \emph{interpretability} as the ability to provide a \emph{cause}, expressed in terms of original data attributes, for an observed \emph{effect}, i.e., attribution of data instance to a certain group. In other words, we are looking for interpretable learning methods to find explanations for class-uniform subsets of labelled data.

Among the variety of machine learning techniques and algorithms, decision trees~\cite{Safavian1991,Breslow97} stand out as the most intuitive method for classification, prediction, and facilitation of decision making. A classification tree consists of three types of nodes: \emph{root node} is the top node comprising all the data, \emph{splitting node} is a node that assigns data to a subgroup, and \emph{leaf} or \emph{terminal node} that represents the final decision, e.g., the outcome of the classification. Importantly, the decision tree can be linearized into decision rules, where the outcome is the contents of the leaf node, and the conditions along the path form a conjunction in the \emph{if-clause}:

{\small
$$\mbox{if }condition_1 \mbox{ and } condition_2 \mbox{ ... and } condition_n \mbox{ then } outcome.$$}
This property highlights decision trees as the most prominent approach for our application where explaining the outcomes of data analysis is the key requirement.

Most commonly decision trees are used for  classification tasks. In such tasks, samples are submitted to a test in each node of the trained decision tree and guided through
it based on the result. However, with few simple adaptations, decision trees can also be used for
clustering~\cite{Castin2018,Liu2005}. \emph{Clustering} is an important exploratory data analysis task in which objects are organized into similarity groups or clusters. In such applications, special split criteria are used to construct the tree without the knowledge of sample labels. The obtained sub-clusters, represented by leaf nodes, are then merged into actual clusters. 

Clustering is often called \emph{unsupervised} learning as no classes denoting an a-priori partition of the objects are known. In contrast, \emph{supervised} learning deals with data records labeled with known classes. In this regard, our problem is similar to the problem of \emph{supervised clustering}~\cite{Eick04} since we are interested in partitioning labelled data into class-uniform regions.

Existing methods for supervised clustering are predominantly density-based methods~\cite{Kriegel11} which imply metrics~\cite{Yu06} to compute distance between pairs of feature vectors. Real data sets often contain symbolic data with no natural order. High data space dimensionality also brings significant challenges for practical application of distance-based clustering~\cite{Jahirabadkar13}. Moreover, such methods do not guarantee that discovered clusters will be explainable~\cite{Roscher20}.

In this paper, we propose a practical method for cluster extraction from a labelled dataset using decision trees in which nodes accumulate large groups of class-uniform instances. Assuming no prior knowledge about the labelled dataset, we create a universal data processing pipeline and node selection criteria to improve the ability of decision trees to locate best quality groups. 
After linearization, the path to the selected nodes in the decision trees provide rules for the generation of comprehensible textual explanations of extracted groups.

The rest of the paper is organized as follows. In Section~\ref{sect:Method}, we discuss the basics of the decision tree-based classification method at the core of our supervised clustering method. In Section~\ref{sect:Preprocessing}, we describe the proposed data preparation pipeline. In Section~\ref{sect:Validation}, we demonstrate the ability of our approach to identify clusters of dense data in a sample dataset. Section~\ref{sect:Stability} presents a method for evaluating stability of extracted clusters. In Section~\ref{sect:RelatedWork}, we overview related work.  Finally, in Section~\ref{sect:Conclusions}, we draw conclusions and outline future work.     

\section{Searching for clusters using decision trees}
\label{sect:Method}

To separate data in different classes, we use a classical CART-like (Classification and Regression Tree) implementation~\cite{Breiman84,Gulati16} of the decision tree which distinguishes nominal and ordinal data and aims at optimizing a given impurity metric such as Gini index or entropy~\cite{Mussard03}. 

The basic logic of the algorithm's training loop is outlined in Listing~\ref{alg:DT}. For every value of every attribute in the data space, the algorithm computes the gain yielded by splitting the dataset into two classes with respect to the pivot value, and remembers the split with the largest gain. If a split with positive gain (or a gain exceeding a given threshold) is found, the splitting node is added to the decision tree, and the cycle is repeated for the left and right partitions of the split dataset until the impurity-decreasing split is no longer possible. Terminal nodes are then assigned classification labels by the majority of data labels that fall into the leaf node. The algorithm works both for binary and multi-label classification. 

\begin{lstlisting}[caption={Decision tree best split evaluation loop}, mathescape=true, label={alg:DT}, language={pseudo}]
for each attribute $A_i \in D: \{A_1, A_2,..., A_d\}$ do
  /*Initialize splitting node*/
  $S \leftarrow \{0, \_, \_, \_, \_\}$ 
  for each pivot $p \in A_i$ do
    if ($A_i$ is ordinal) then
      $D_1 \leftarrow \{d \in D \, | \, x_j \leq p\}, \, D_2 \leftarrow \{d \in D \, | \, x_j > p\}, \, x_j \in A_i$ 
    else 
      $D_1 \leftarrow \{d \in D \, | \, x_j = p\}, \, D_2 \leftarrow \{d \in D \, | \, x_j \neq p\}, \, x_j \in A_i$ 
    endif  
    $g \leftarrow gain(D_1, D_2)$
    if (S.g < g) then
       /*Gain, pivot, attribute, two partitions*/
       $S \leftarrow \{g, p, A_i, D_1, D_2\}$
    endif   
  endfor
  if (S.g > 0) then
    /*add splitting node S to the decision tree*/
    $T \leftarrow S$ 
  endif   
endfor
\end{lstlisting}

The decision tree construction algorithm recursively runs over all predictors and for each unique value, called a pivot, computes the impurity metrics based on uniformity of data in each split. The split condition depends on the type of the data column:
\begin{itemize}
    \item for \emph{ordinal} data (e.g., numeric, date/time or ordinal symbolic columns), we split the dataset depending on whether the column value in each row is less than or equal to, or greater than the pivot.
    \item for \emph{nominal} data (nominal symbolic or Boolean columns, we split the dataset depending on whether the column value in each row is equal or not equal to the pivot.  
\end{itemize}
The attribute values can be replaced with ordinal numbers corresponding to unique values in the original attribute, preserving their natural order or establishing an artificial order, i.e., we do not actually look at the original feature values in the impurity-decrease evaluation loop.

The decision tree continues branching until dataset purity cannot be further improved or any of the early pruning conditions (tree depth, minimal impurity decrease, etc.) are satisfied. Each new split improves overall classification accuracy but contributes to a larger set of conditions (terminal nodes) reachable via longer paths, and each node covers a smaller portion of the training dataset. The early pruning options may prevent the decision tree classifier from over-fitting, but it is not clear how to set parameters that would guarantee the construction of the best possible decision tree for the purpose of supervised clustering of an unknown  dataset. Hence, we need a method to rank nodes with various levels of impurity and sample sizes in order to choose the most significant groups of vectors of a given class.

For example, the decision tree classifier below splits sets of 887 Titanic passengers into two classes: those who survived in the disaster $(decision = 1)$ vs those who did not $(decision = 0)$. Our goal is to characterize the sets of survivors in as few terms as possible. 

To determine which nodes form the best cluster candidates, we rank them by the combination of precision and recall using the F-measure: 
$$F_{\beta} = (1 + \beta^2)\frac{precision + recall}{\beta^2 \,precision + recall},$$ $$precision = \frac{ tp}{tp+fp}, \, recall = \frac{tp}{tp+fn}, $$ where $tp$ (true-positives) is the number of correctly classified entries in the node, $fp$ (false-positives) is the number of wrongly classified in the node, and $fn$ (false-negatives) is the number of positives not in the node. The parameter $\beta$ is chosen in such a way that recall is considered $\beta$ times as important as precision.

\begin{figure*}
    \centering
    \subfigure[F-0.5]{
         \includegraphics[width=0.9\textwidth]{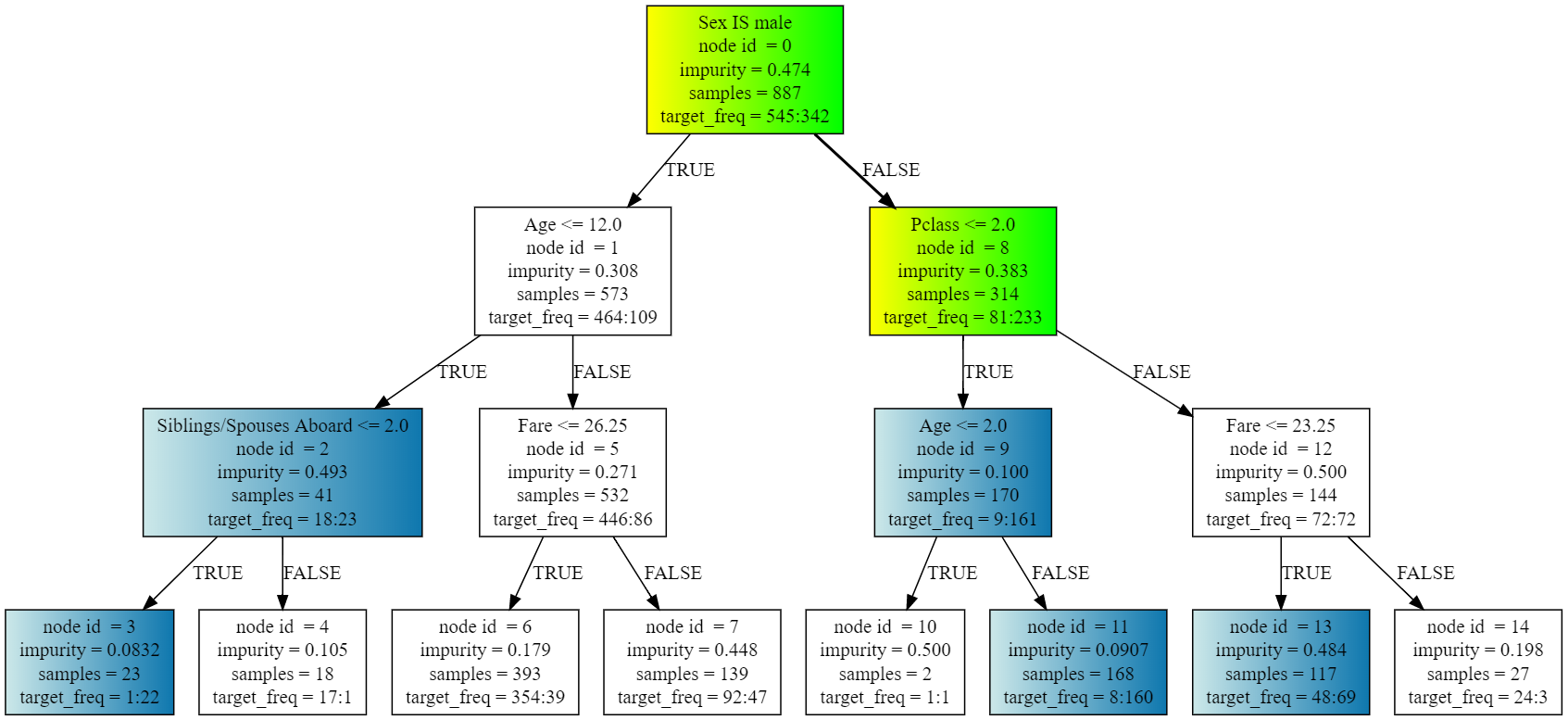}
         \label{fig:ds-titanic-tree1}
    }
    \subfigure[F-0.33]{
         \includegraphics[width=0.9\textwidth]{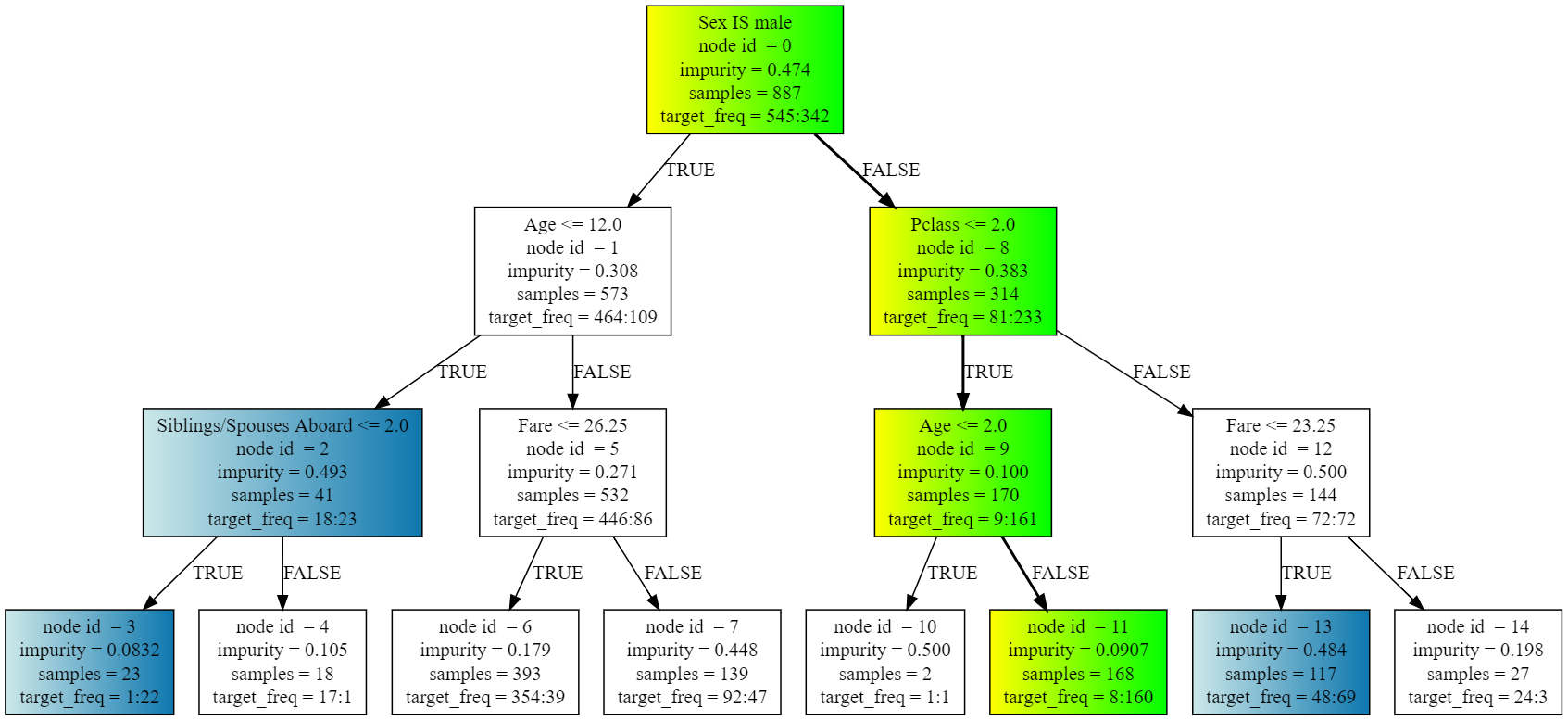}
         \label{fig:ds-titanic-tree2}
    }
    \caption{Extracting clusters from a decision tree using $F_\beta$, \emph{Titanic} dataset}
    \label{fig:ds-titanic}
\end{figure*}

\setlength{\tabcolsep}{0.5em}
\begin{table*}
    \centering
    \caption{Metrics for cluster extraction}
    \label{tab:F-beta}
    \begin{tabular}{|c|c|c|c|c|c|c|}
        \hline
        {\bf Cluster} & {\bf Gini impurity} & {\bf Size} & {\bf Precision} & {\bf Recall} & {\bf F1} & {\bf F-0.5} \\
        \hline
         $c_1$ & 0.3828 & 314 &	 0.7420 & 0.68128 &	0.71037 & 0.7290 \\
         $c_2$ & 0.0907 & 168 &	0.9523  & 0.46784 &	0.62745 & 0.7889 \\
         $c_3$ & 0.1002 & 170 &	0.94706	& 0.47076 &	0.62891 & 0.7876 \\
         $c_4$ & 0.0832 & 23  &	0.95652	& 0.06433 &	0.12054 & 0.2534 \\
         $c_5$ & 0.4839 & 117 &	0.5897	& 0.20175 &	0.30065 & 0.4259 \\
         $c_6$ & 0.4925 & 41  &	0.5610	& 0.06725 &	0.12010 & 0.2272 \\
         \hline
    \end{tabular}
\end{table*}

For the Titanic dataset, using F1 for node ranking would yield node $n_8$ (see Figure~\ref{fig:ds-titanic-tree1}) selected as the best cluster group. For F-0.33, the node $n_{11}$ in Figure ~\ref{fig:ds-titanic-tree2} provides the best score. Parameter $\beta$ defines user preference for more pure vs larger group of data in the target class.

If more groups need to be extracted from the same tree, we select nodes with the highest metric which are not ancestors or descendants of previously selected nodes. Thus, for the tree above, two clusters would be identified: the node $c_1$ (which excludes $c_2$, $c_3$ and $c_5$ as its descendants) and $c_4$ (which excludes $c_6$ as its ancestor). However, we show later that picking one cluster from the tree and retraining the classifier to find the next best node is a better strategy as it allows us to produce subsequent clusters with higher-density.  

\section{Data preprocessing pipeline}
\label{sect:Preprocessing}

For the suggested clustering method, it is important to build shallow decision-tree based classifiers with reasonable performance. Below we describe helpful data pre-processing steps to enable decision trees to locate larger clusters of a given class using a limited number of test nodes.

\subsection{Binning} 
Data \emph{binning} (also called discrete binning or bucketing) is a data pre-processing technique used to reduce the effects of minor observation errors. The original data values which fall into a given small interval, a bin, are replaced by a value representative of that interval (e.g., the central value for numeric data). 

Binning helps to speed up the decision tree construction by reducing the number of unique values or pivot points for tree branching. Among popular methods are \emph{equal-width binning} and \emph{quantile-based binning}. The former accepts as input the number of bins, and using the min/max values of the dataset, defines the bin width. The latter uses quantiles (median, quartiles, deciles, percentiles, etc.)  to determine the number and width of bins. In our project, we apply the percentile binning to transform the numeric features before clustering.

An excessive number of unique values in symbolic features negatively affects the performance of the decision tree since the algorithm takes time to iterate over all such values and evaluate dataset partitions with respect to the pivot. Hence, we introduced three methods to group symbolic values into a given number of bins and construct the decision tree by splitting entries based on the new label representing the set of grouped original values:

\begin{itemize}
    \item 
In the most general case, symbolic values are grouped together into sets of equal width. This procedure does not help to improve classification accuracy, but speeds up the decision tree construction by reducing the number of unique symbolic values.   
    \item
Alternatively, we may order the column unique values by the frequency of their occurrence in the dataset, and then split them into a given number of bins. Frequently used ordinals may remain unchanged or will be grouped together with just a few other values while rarely occurring ordinals are going to be grouped together in such a way that their joint occurrence frequency approaches the estimated mean value. This method is beneficial for datasets with unbalanced feature values and helps to produce more intuitive explanations of the extracted groups by reverting the path clause, e.g., ``customer is \emph{not} in the \emph{United States}'' as opposed to ``customer is in \{long list of countries\}''.
    \item
    Yet another binning method is based on symbolic value similarity: values sorted in lexicographic order are split into a given number of bins. Such transformation is most suitable for hierarchical data such as postcodes or phone numbers where initial symbols represent a meaningful category (e.g., geographic region). The categorical bins may either include an equal number of ordinals (equal-width binning) or rely on a syntactic similarity metric to automatically group most similar categories into the same bin. For example, in our implementation, we use the \emph{Jaro-Winkler distance}~\cite{Winkler1994,jaro95} between $i^{th}$ and $(i+1)^{th}$ values on a sorted set of unique symbolic values and select ordinals with the largest distance. The similarity-based transformation produces the required number of bins of different length; the values in the same bin are likely to have a common prefix. 
\end{itemize}

Customer datasets often include date/time related information such as e.g., operation timestamp, contract start date, card renewal date, time of customer call, etc. Such features typically have many unique values and will be useless for machine learning without adequate processing. Moreover, the huge number of unique values affects the speed of the CART algorithm. Hence recognizing date/time features among other columns followed by sorting and binning of their values is an essential operation in our cluster-extraction pipeline.

Sorted date/time values can be binned according to two types of interval-forming strategies:
\begin{itemize}
    \item \emph{Frequency-based} intervals: sorted date, time or date/time value column is split to the given number of varying-length bins so that each bin covers an interval with approximately equal number of values in the dataset.
    \item \emph{Equal-width} intervals: sorted date, time or date/time column is split into approximately equal-length intervals based on the smallest and largest values in the dataset.
\end{itemize}

\subsection{Ordinal encoding}
While values in numeric columns can be compared by their natural order (with an exception of rows with missing values), comparing symbolic data is subject to interpretation. Categorical data can be treated as nominal or ordinal. In nominal columns, data is not arranged into any kind of order by their position, i.e., we do not imply that in a set 

{\small
$$S = \{1: Amsterdam, 2: London, 3: New York, 4: Shanghai\}$$ 
}
it holds that 

{\small
$$Amsterdam \leq London \leq New York \leq Shanghai.$$
} 
Alternatively, we can artificially impose a partial order on this set, e.g., implying that the cities are arranged by their size or alphabetic order. 

When comparing symbolic column data to a pivot, we use equality operator for nominal columns and less-or-equal-than operator for ordinal columns, applied to their position (ordinal) in a set of unique values in this column. For example, valid split conditions for the aforementioned nominal set $S$ are 

{\small
$$(x = London) \mbox{ vs } ((x \neq London) =$$ $$\{Amsterdam, New York, Shanghai\}),$$
}
while valid conditions for the same set treated as ordinal are 

{\small
$$((x \leq London) = \{Amsterdam, London\}) \mbox{ vs } $$ 
$$ ((x > London) = \{New York, Shanghai\}).$$
}
It is easy to observe that the partial order on symbolic ordinal data can significantly affect the ability of the decision tree to find an optimal split to minimize data impurity in the column splits. For example, no single split of the ordinal set above will produce subsets 

{\small
$$\{Amsterdam, New York\}\mbox{ and }\{London, Shanghai\}.$$}
Hence, it is preferable to map symbolic data to the set of ordinals (or equivalently, sort the set of unique symbolic values in a column) in such a way that there exists a pivot $p$ such that the column split producing sets $\{ x \leq  p\}$ and $\{x > p\}$ minimizes the impurity metric. For  labelled datasets with binary targets, this condition can be met by sorting of column unique values by frequency occurrence in one of the target classes. For example, for a labelled dataset with an attribute \emph{city} with ordinal values in the ordered set $S$ shown above and a the \emph{class} attribute supplying a label for each entry
\begin{equation*}
A_{city, class} = 
\begin{pmatrix}
1 & 2 & 3 & 4 & 4 & 3 & 2 & 1 \\
1 & 0 & 1 & 1 & 0 & 1 & 0 & 1
\end{pmatrix}
\end{equation*}
the contingency table is shown in Table~\ref{tab:contingencyTable}.
After sorting the unique values by frequency occurrence in the target group 0, the following ordinal mapping will be created: 

{\small
$$\{1: London, 2: Shanghai, 3: Amsterdam, 4: New York\}.$$} The CART-like algorithm, iterating over feature ordinals, will find out that selecting \emph{Shanghai} as a pivot generates splits
{\small $$s_1 = \{London, Shanghai\}$$} and 
{\small $$s_2 = \{Amsterdam, New York\},$$}
corresponding to labels $[0,1,0,0]$ and $[1,1,1,1],$ which yield the minimal impurity for this feature. 
\begin{table}
    \centering
    \caption{Contingency table: occurrence frequency in a given class}
    \label{tab:contingencyTable}
    \begin{tabular}{|c|l|c|c|}
        \hline
        {\bf Ordinal} & {\bf Category} & {\bf Class=0} & {\bf Class $\neq$ 0}\\
        \hline
        1 & Amsterdam & 0/2 & 2/2 \\
        2 & London    &	2/2 & 0/2 \\
        3 & New York  &	0/2	& 2/2 \\
        4 & Shanghai  &	1/2	& 1/2 \\
        \hline
    \end{tabular}
\end{table}

With many unique values in a column, equality-based splitting of nominal data for the construction of a shallow decision tree (with depth up to 5 tree levels) is unlikely to reveal large opportunity groups. Symbolic sorting by frequency increases these chances by placing all values that correlate with a given target class on one side of the split-able ordinal set. 

Features with too many unique values should either be binned using frequency or similarity-based binning methods or excluded from the dataset as we either will not be able to generalize observations based on these features, or risk to define clusters described by the condition $x\in T,$ where $T$ is a large set of original feature values produced by the symbolic binner.  

\subsection{Best node selection}

Listing~\ref{alg:CL} outlines the process of extracting class-uniform clusters defined by nodes in each decision tree classifier. The ranking metric $F_{\beta}(S, \beta, c)$ evaluates the F-measure on the data in node $S$ for the given value of $\beta$ and decision class $c$. The node with the highest $F_{\beta}$ defines the best cluster.
Generally, this class may not coincide with the predicted tree node label defined by the majority class of the entries in the node. In other words, the method may also be used to discover ``weak clusters'' in the required class, i.e., those with elevated density over the whole population but with smaller number of entries than in other classes. 

\begin{lstlisting}[caption={Cluster extraction  algorithm}, mathescape=true, label={alg:CL}, language={pseudo}]
  $S_{best} \leftarrow \{\}$
  $D \leftarrow \{A_1, A_2,... A_d\}$
  for k: 1..N do
    train classifier $T_k$ on dataset $D$
    $F_k = 0 \quad S_k \leftarrow null$
    for each node S in $T_k$ do
      $F_S \leftarrow F_{\beta}(S, \beta, c)$
      if ($F_S > F_k$) then
        $F_k = F_S \quad S_k \leftarrow S$
      endif    
    endfor    
    if $S_k \neq null$ then
      $S_{best} \leftarrow S_{best} \cup S_k \quad D \leftarrow D \setminus S_k.D$
    endif  
  endfor
\end{lstlisting}

Along with data preprocessing methods, it is useful to include a \emph{node removal} operation to our pipeline to exclude data from the best cluster discovered by the $i$-th decision tree before training the next classifier. 

\section{Validation}
\label{sect:Validation}

In this section, we validate our dense class-uniform group extracting method using a dataset with artificially created groups of customers interested in an imaginary proposition. The customer description in this case study is based on the \emph{Adult} (also known as \emph{Census Income}) dataset~\cite{Dua:2019} from the UCI Machine Learning repository. The (train part) of the original dataset consists of 32561 rows in 15 columns. We kept the original fields and added a new class label that indicates whether the user is interested in a proposition. We also altered selected attribute values to add hidden higher-density groups of users into this dataset.

The hidden groups we synthesized within this dataset can be identified using the following conditions:

{\small
\begin{enumerate}
    \item fnlwgt $\leq  285194.62$ \& native-country $\in$ \emph{\{Ecuador, El-Salvador, Haiti, Cuba, France, Yugoslavia, Germany, ? (unknown), Poland, Hungary, Laos, Mexico, Japan, Hong, Vietnam, Peru, England, United-States\}} - 17.8\%,
    \item occupation = \emph{Exec-managerial} \&  capital-gain $\leq -75.82$ \& race $\in$ \emph{\{Amer-Indian-Eskimo, Asian-Pac-Islander, White\}} - 5\%,
	\item capital-loss $ \leq 115.42$ \& education-num $\leq 9.1$ \& income = ``$>50K$'' - 4.2\%,
	\item hours-per-week $\leq 35.12$ \& marital-status $\in$ \emph{\{Widowed, Married-spouse-absent, Divorced\}} \& relationship = \emph{Not-in-family} - 1.6\%.
\end{enumerate}
}
For the sake of presentation, we trained decision trees with 3 levels. In practice, deeper decision trees can be constructed as our $F_\beta$-based node ranking algorithm evaluates all nodes in the tree, not just terminal nodes. This makes the proposed  cluster-extraction method immune to classification tree over-fitting. 

We are interested in nodes of the right class, the smallest impurity and the largest samples size. According to these criteria (estimated by $F_\beta$ with $\beta=0.33$) the best node in the tree is the left most leaf 

\begin{figure*}
    \centering
    \subfigure[Group 1]{
         \includegraphics[width=0.9\textwidth]{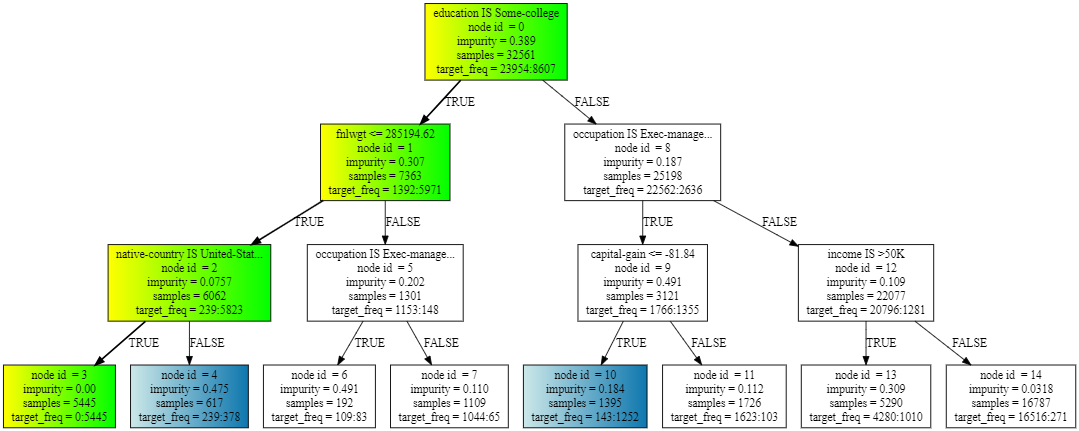}
         \label{fig:ds-adult-1}
    }
    \subfigure[Group 2]{
         \includegraphics[width=0.9\textwidth]{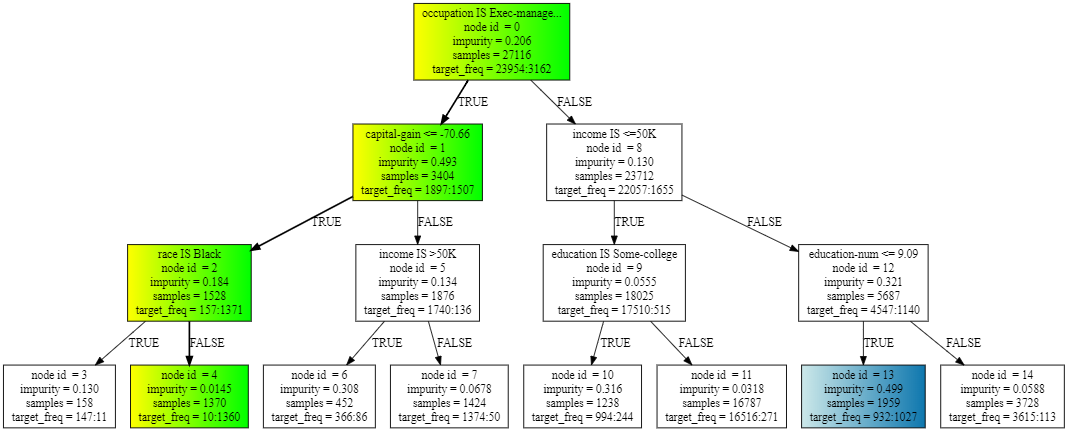}
         \label{fig:ds-adult-2}
    }
    \subfigure[Group 3]{
         \includegraphics[width=0.9\textwidth]{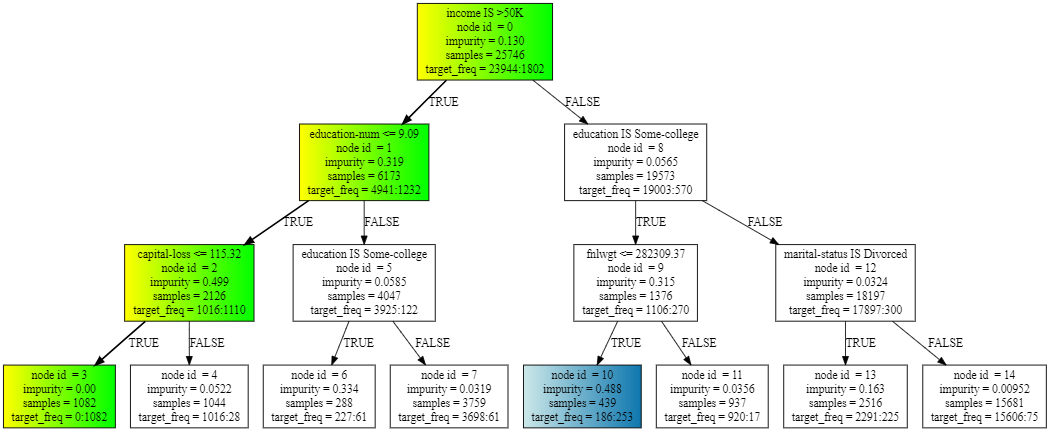}
         \label{fig:ds-adult-3}
    }
    \caption{Extracting clusters from a decision tree, synthetic \emph{Adult} dataset}
    \label{fig:ds-adult}
\end{figure*}

{\small
$$n_3 =(id=3, decision=1, impurity = 0, samples = 5445)$$}
highlighted in Figure~\ref{fig:ds-adult-1}. The path to this node reveals 2 conditions involved into the definition of the group 1. The cluster covers 16.7\% of the whole population while the whole hidden group covers 17.8\%.  

It is easy to see the similarity of the path conditions in the other two trees with properties defining the remaining hidden groups. Note that the 2nd best group, identified by the path to the highlighted node $n_4$ in Figure~\ref{fig:ds-adult-2}, contains more entries of the required class (1360) than the second best node in the first tree (node $n_{10}$, 1252 entries). Similarly, the third group shown in Figure~\ref{fig:ds-adult-3} quality benefits from the node removal and tree retraining (1082 entries vs 1027 entries).

\begin{figure*}
    \centering
    \includegraphics[width=\textwidth]{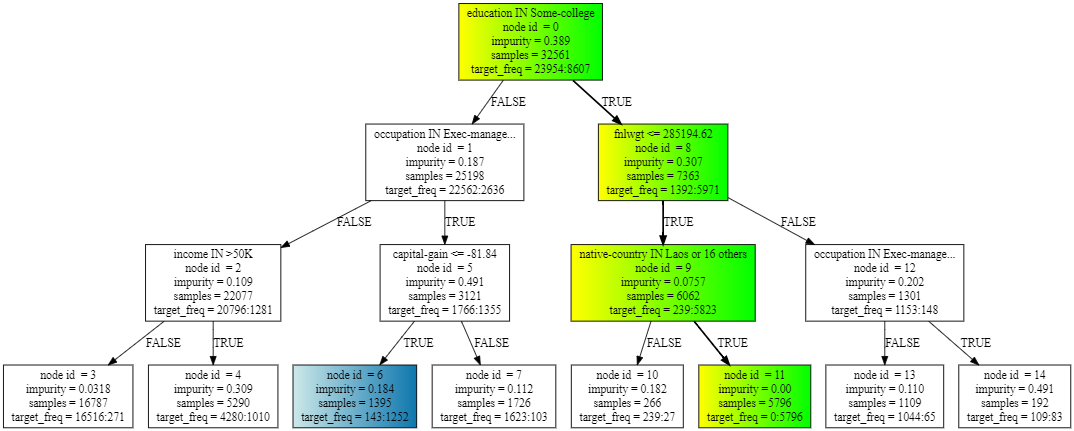}
    \caption{Extracting clusters from a decision tree using F-Beta score, Synthetic adult dataset with symbolic reordering}
    \label{fig:ds-adult-reordered}
\end{figure*}

To produce an optimal split in the decision tree algorithm, i.e., the split that minimizes the given impurity metric, we need to compare symbolic values in each column to a chosen pivot in this column. Figure~\ref{fig:ds-adult-reordered} shows the effect of symbolic data sorting on supervised cluster extraction: this tree identifies the largest group of customers corresponding to its leaf node 

{\small
$$n_5 = (id=11, decision=1, impurity=0.0, samples=5796)$$} 
with the following conditions:
{\small
$$\mbox{education} > 12th \,\, \& \,\, \mbox{fnlwgt} \leq 285194.62 \,\, \& $$
$$\mbox{native-country} > Domenican\mbox{-}Republic.$$}

The unique values of variable \emph{education} are ordered as follows: 
\vspace{1ex}\\
\fbox{
    \begin{minipage}{0.45\textwidth}
    \centering
    \scriptsize 
    \emph{\{1st-4th, 7th-8th, Prof-school, HS-grad, Bachelors, 5th-6th, Doctorate, 11th, 10th, Masters, Preschool, Assoc-acdm, Assoc-voc, 9th, 12th, Some-college\}}.
    \end{minipage}
}
\vspace{0.5ex}\\
Hence, the following conditions are equivalent:
{\small 
$$\mbox{education} >12th \equiv \mbox{ education } \in \{\mbox{ \emph{Some-college}}\}.$$}
The \emph{native-country} values re-ordered by class-uniform frequency occurrence  \vspace{1ex}\\
\fbox{
    \begin{minipage}{0.45\textwidth}
    \centering
    \scriptsize 
        \emph{\{Holand-Netherlands, Trinadad\&Tobago, Italy, Nicaragua, Portugal, Scotland, Outlying-US(Guam-USVI-etc), Thailand, China, France, Columbia, Canada, Philippines, South, Iran, India, Greece, Cambodia, Puerto-Rico, Taiwan, Guatemala, Honduras, Ireland, Jamaica, Dominican-Republic, Laos, Cuba, ?, Germany, Hong, Yugoslavia, Mexico, United-States, England, Peru, Poland, El-Salvador, Haiti, Japan, Vietnam, Hungary, Ecuador\}.} 
    \end{minipage}
}
\vspace{0.5ex}\\
yields the decision tree split 
{\small
$$\mbox{native-country } > \mbox{Domenican-Republic}$$} 
corresponding to the set of 17 countries that follow the pivot value in this ordered set, including \emph{United-States} used in the description of the hidden group 1. This cluster retrieves the whole 17.8\% of the population and covers 100\% the main hidden group, demonstrating that symbolic ordering improves the overall performance of the proposed cluster extraction method.  

As this case study shows, the method we designed is able to reveal hidden trends in labelled datasets with high accuracy and high coverage.

The method's execution time is comparable to the execution time of the decision trees on typical classification tasks, multiplied by the number of selected cluster groups. Some overhead can be added by pre-processing steps, i.e., binning or symbolic ordering. 
The experiment with the \emph{Adult} dataset described in this chapter that applies symbolic field reordering, selects 3 best clusters, evaluates performance and visualizes corresponding decision trees is performed within 4 seconds. Our implementation is based on the Python's Scikit-Learn library~\cite{pedregosa2011scikit} and is executed on an ordinary laptop with Intel Core i7 processor and 16GB RAM.  

\section{Evaluating stability of clusters}
\label{sect:Stability}

At the core of the proposed clustering algorithm is a single shallow decision tree.  The paths to decision tree nodes with the largest $F_\beta$ provide us with easy-to-describe class-uniform clusters. However, decision trees are generally considered to be weak learners, i.e., their performance (e.g., prediction or classification accuracy) and stability (preservation of structure, or, most importantly for us, branching conditions on paths to nodes defining class-uniform clusters, in response to small changes in the input data) can be rather poor. More accurate ensemble methods such as random forests~\cite{Breiman01} or gradient boosting~\cite{Friedman02} are often used in practice for prediction and classification tasks, but we are bound to decision trees for the sake of interpretability. While precision and recall characterize the size and purity of the identified clusters, their stability remains unclear. Therefore, we drafted a simple method to estimate stability of discovered clusters using the bagging technique. 

We generate $N$ samples from the original dataset. For each sample, we can compute extracted cluster (decision tree node) stability score as follows: let $c(D)$ be a set of entries from the original dataset $D$ that fall into cluster $c$, and $c'(D_k)$ be a set of entries from a sample dataset $D_k \subseteq D$ that fall into an extracted cluster $c'$, the stability score between $c(D)$ and $c'(D_k)$ is computed as: 

{\scriptsize
$$S(c(D), c'(D_k)) = \frac{| c(D) \cap  c'(D_k) \cap D_k|} {|c(D) \cup c'(D_k) \cap D_k|} = \frac{| c(D_k) \cap  c'(D_k)|} {|c(D_k) \cup c'(D_k)|},$$}
i.e., we measure what fraction of the population included to the sample dataset $D_k$ is included both into $c$ and $c'$. 

We compare each cluster of the original dataset with each cluster of the sample dataset. This is useful because they may be ordered differently, e.g., cluster $c_i$ may resembles more, in terms of expressions in the conjunction of if-clauses, cluster $c'_j$ of the sample dataset rather than $c'_i,$ where $i = 1..n$ represents place in the ranked list of clusters. Hence, the overall stability metric of $c_i$ can be computed as maximum among its pairwise scores with each and every cluster from the sample dataset $D_k$: 
$$S_k(c_i) = \max_{j=1..n}S(c_i(D), c'_j(D_k)).$$ 
The stability score of the group $c_i$ is an average score among all samples: 
$$ S(c_i) = \frac{1}{N}\sum_{k=1}^{N}S_k(c_i).$$ 

Using this method, we estimated that the main cluster extracted from the \emph{Titanic} dataset exhibits stability score around 90-98\%, i.e., small variation in the input data does not affect the decision rules for our class-uniform clustering. In the contrary, for the \emph{Adult} dataset, the best cluster is rather unstable with the stability score around 20-25\%, indicating that discovered clustering rules may not provide the best result when applied to a larger population of similar data. 

\section{Related work}
\label{sect:RelatedWork}
Machine learning (ML) methods have been remarkably successful for a wide range of application areas in the extraction of essential information from data. In many applications, however, we cannot rely on the automated decision-making without understanding all the circumstances around them. Three core elements of ML: transparency, interpretability, and explainability, play a crucial role in adaptation of ML methods in critical domains. Roscher et al.~\cite{Roscher20} provide a survey of recent scientific works that incorporate explainable ML. In our application, the aforementioned characteristics are key requirements, which largely predetermined the use of decision trees for cluster extraction.

Decision trees divide an input space into  smaller regions and make prediction depending on a region. The state-of-the-art decision-tree-based ensemble models such as random forests~\cite{Breiman01} and boosted trees~\cite{Friedman02,XGBoost} are extremely popular ML  methods because of their high predictive performance.
Such methods, commonly known as additive-tree-models, generate a large number of regions, which roughly means that there are thousands of different rules for prediction. People, interested not only in prediction outcome but also in the reasons why such a prediction was made, cannot understand such tremendous number of rules. 

A single decision tree~\cite{Breiman84} is one of the most interpretable models. However, while its predictive power is rather weak, the number of regions generated by a single tree may still be significant. There are continuous efforts to control the shape of the decision trees via regularization~\cite{Scheffer00} or feature engineering~\cite{Deng12}.

Eick et al.~\cite{Eick04} recognised the need for supervised clustering. They introduced four representative-based algorithms for supervised clustering: (i) a greedy algorithm with random restart that seeks for solutions by inserting and removing single objects from the current solution, (ii) SPAM, a variation of the Partitioning Around Medoids (PAM) clustering algorithm, (iii) an evolutionary computing algorithm, and a (iv) fast medoid-based Top-Down Splitting (TDS) algorithm. 
These methods rely on distance metrics and thus do not provide easy interpretability. It may be interesting to explore the combination of medoid-based clustering algorithms with decision explanation techniques in the context of our application. 

Liu et al.~\cite{Liu2005} train the decision tree to partition unlabelled data space into clusters. In contrast, our approach works with labelled data. By removing data from the best tree node and retraining the tree, we find the next best cluster with potentially better performance metric than any split outlined by a single tree. 

In another approach to clustering with decision trees~\cite{Castin2018}, a divisive agglomerative approach is proposed. Two split criteria, box volume and graph closeness, are used to identify cluster boundaries, applied to each feature separately. While graph closeness is a distance-based metric with the disadvantages this entails, the former metric starts from lower bound and upper bounds placed around the class mean. Then, the number of samples with feature value in between the two bounds are counted. Step by step, the bounds are moved apart and the samples counted again. When 95\% of the samples are captured, the width of the box is defined for that dimension, the box volume is the product of all widths. 

There are numerous works dedicated to the analysis of stability of ML methods, including interpretable models~\cite{Interpretable-stability} and stability of clusters~\cite{Cluster-stability}. We adopted an approach based on region compatibility~\cite{Region-stability}. In contrast to this work, our method measures stability not of the entire tree but of partitions representing class-uniform clusters on bagged datasets.   
\section{Conclusions and Future Work}
\label{sect:Conclusions}
We discussed the problem of discovering dense class-uniform regions in labelled datasets. Our solution relies on iterative training of decision tree classifiers with a posteriori node ranking method to select largest and most pure clusters. 

As future work, we plan to explore other ways to locate interpretable regions of class-uniform data. We are also working on a corpus to evaluate such algorithms. Yet another line of work in the context of the existing application involves pre-processing pipeline customization using fast criteria to decide which optimization techniques, binning and feature extraction methods should be applied to various types and distributions of attributes in online datasets.

\section*{Conflict of interest}
{\small
On behalf of all authors, the corresponding author states that there is no conflict of interest.
}

\bibliographystyle{spmpsci}
\bibliography{main}

\end{document}